\documentclass[letterpaper, 10 pt, conference]{ieeeconf}

\IEEEoverridecommandlockouts
\usepackage{cite}

\usepackage[pagewise]{lineno}
\usepackage[ruled,vlined]{algorithm2e}
\usepackage{amsmath}
\usepackage{amssymb}
\SetKwComment{Comment}{/* }{ */}
\usepackage{multirow}

\usepackage{bbold}
\usepackage{graphicx}
\usepackage{subcaption}
\usepackage{xcolor}

\graphicspath{{./figures/}}

\title{\LARGE \bf
Towards Optimal Head-to-head Autonomous Racing with Curriculum Reinforcement Learning 
}
\author{Dvij Kalaria$^{1}$, Qin Lin$^{2}$, and John M. Dolan$^{1}$
\thanks{$^{1}$The authors are with the Robotics Institute, Carnegie Mellon University {\tt\small dkalaria@andrew.cmu.edu},{\tt\small jdolan@andrew.cmu.edu}}
\thanks{$^{2}$Qin Lin is with the Electrical Engineering and Computer Science Department, Cleveland State University {\tt\small q.lin80@csuohio.edu}}
}

\begin{document}

\maketitle
\thispagestyle{empty}
\pagestyle{empty}

\begin{abstract}

Head-to-head autonomous racing is a challenging problem, as the vehicle needs to operate at the friction or handling limits in order to achieve minimum lap times while also actively looking for strategies to overtake/stay ahead of the opponent. In this work we propose a head-to-head racing environment for reinforcement learning which accurately models vehicle dynamics. Some previous works have tried learning a policy directly in the complex vehicle dynamics environment but have failed to learn an optimal policy. In this work, we propose a curriculum learning-based framework by transitioning from a simpler vehicle model to a more complex real environment to teach the reinforcement learning agent a policy closer to the optimal policy. We also propose a control barrier function-based safe reinforcement learning algorithm to enforce the safety of the agent in a more effective way while not compromising on optimality.

\end{abstract}


\section{Introduction}

There has been a growing interest in autonomous racing research in recent years \cite{Betz2022AutonomousVO} also accelerated by competitions such as RoboRace \cite{Rieber2004TheRC}, F1Tenth \cite{OKelly2019F1TENTHAO}, and the Indy Autonomous Challenge \cite{Wischnewski2022IndyAC}. Professional human race drivers operate to follow racing lines to achieve optimal performance and tend to outperform opponents while adhering to the racing rules. Prior works in autonomous racing tend to ignore the latter and only consider collision avoidance. It is difficult to inculcate these complex rules and design a classical rule-based controller which takes care of all scenarios and tackles a wide range of opponent behaviors. Most previous reinforcement learning (RL)-based works tackling this do not include racing line information in the framework \cite{Thakkar2022HierarchicalCF} \cite{Herman2021LearntoRaceAM}, due to which it becomes difficult for the learning agent to learn an optimal policy which can be generalized to other tracks. Also, training an agent directly on a complex racing environment with complex vehicle dynamics debars the learning agent from learning complex behaviors like skidding. In this work, we propose a complex racing environment which can be used to train head-to-head RL agents to learn an optimal policy for competing with the opponents. Then, we propose a curriculum learning-based framework which transitions the vehicle model from a simple to a more complex one to tackle the problem of this sub-optimality. Some works \cite{Cheng2019EndtoEndSR} also use Control Barrier Functions (CBFs) as a shield while learning, which has also been proven to boost the safety performance. However, these safety constraints may debar the agent from learning a high-performant policy, especially in competitive environments like ours where performance is also a primary concern alongside safety. In this work, we also propose a curriculum learning-based CBF framework to enforce safety during learning while not compromising on the optimality of the final learned policy. We gradually remove the CBF interference with the policy as the agent progressively learns to be more safe, hence focusing on improving performance next. We test our controllers against baselines by performing head-to-head races with other baselines. We briefly summarize our contribution as follows: 1) We propose a head-to-head racing environment which models complex vehicle dynamics and collision among agents or with walls; 2) We design an effective hierarchical controller which includes racing line information inside the hierarchical controller to train an optimal policy; 3) We design a curriculum learning-based framework which effectively enables learning an optimal policy for the agent.

The rest of the paper is organized as follows: Section \ref{sec:related} briefly discusses the previous related works. Section \ref{sec:preliminaries} presents the problem formulation. Section \ref{sec:proposed} elaborates on the proposed framework. The simulation results are presented in Section \ref{sec:results}. The concluding remarks and future work can be found in Section \ref{sec:conclusion}.


\section{Related Works}
\label{sec:related}
Autonomous racing has received a lot of interest from the research community recently at all levels of the stack including the perception, localization, path planning and control, as discussed in this literature review paper \cite{Betz2022AutonomousVO}. We specifically focus on the path planning and control level of the stack. Most recent works focus on optimizing lap times for a single agent, with very recent works addressing multi-agent planning and control. 

For single-agent, most works propose calculating an optimal racing line offline and using a control algorithm to run on it as a reference online. \cite{Christ2021TimeoptimalTP} propose IPOPT optimization to compute the racing line while \cite{Jain2020ComputingTR} proposes Bayesian optimization to obtain the racing line. \cite{Heilmeier2020MinimumCT} calculates a minimum-curvature path which is very close to the optimal line. \cite{9051390} proposes using an LQR controller to track the racing line. \cite{Curinga2017AutonomousRU} \cite{Alcal2020AutonomousRU} proposes a discrete MPC controller, while \cite{Liniger2015OptimizationbasedAR} proposes a model predictive contouring controller (MPCC). There also have been some recent works to account for model and environment uncertainty \cite{Kalaria2022DelayawareRC} and some \cite{Rosolia2019LearningHT} \cite{Kalaria2023AdaptivePA} to account for model changes online. Several works proposed to use reinforcement learning \cite{Herman2021LearntoRaceAM} \cite{Jaritz2018EndtoEndRD} and imitation learning \cite{Kalaria2023TowardsSA} \cite{Sun2022ABC} to control a vehicle around the track with the objective of minimal race times.

For multi-agent racing control, there are works \cite{Raji2022MotionPA} \cite{Liniger2018PathPA} on using rule-based strategy selection and using high-level path planner/low-level control to execute strategies like overtaking, blocking, collision avoidance, etc. However, these works rely on a lot of parameters and it always difficult to find an optimal set of parameters to work on all track maps and environments. Some works use game theoretic planning followed by classical control \cite{Jung2021GameTheoreticMP}. Some recent works like \cite{Thakkar2022HierarchicalCF} \cite{Francis2022LearntoRaceC2} use reinforcement learning to effectively learn an end-to-end controller that learns to win the race and thus learns certain strategies to do so. The environments used to train the RL agent(s) can be varied with different maps, surfaces, etc. to learn a widely generalizable policy. However, these works still struggle in learning an optimal policy. Some works like \cite{Thakkar2022HierarchicalCF} use a simple kinematic vehicle model to train and test the RL policy with other classical approaches. In this work, we propose a complex racing environment with a dynamic vehicle model with challenging parameters close to the actual racing car to train and test the RL controller. We also propose a curriculum-based course to train the RL agent that helps in learning a better policy. 

\section{Problem Formulation} \label{sec:preliminaries}

We first present our dynamic game formulation. Let there be 2 players $i$ and $j$ racing against each other over $T$ time steps. The track is defined by a sequence of $k$ checkpoints along the center, $\{c_i\}_{i=1}^T$. The objective for each player is to minimize the time difference between it and its opponent in completing the final lap defined as $c_k$. Let $\gamma_i$ be the earliest time step when a player $i$ reaches a checkpoint. Let the state of the vehicle be $x_i \in X \subset R^6$ and control action at each time step be $u_i \in U \subset R^2$. Let $r_i \in {1,2...k}$ be the index of the last checkpoint passed by the player. Let $p : S \rightarrow C$ be a function mapping a state $s$ to a checkpoint. Also, we must ensure the state $s_t$ is always within the track boundaries, i.e., $q(s_t) \le w$ where $q$ is a function for the closest distance to the center line and $w$ is the track width, assumed to be constant. For collision avoidance, let $d : X * X \rightarrow R$ be a function that returns the shortest distance between bodies given the $2$ vehicle states as $d(s_t^i,s_t^j)$. Based on these variables, the objective for agent $i$ is given as:

\begin{equation}
\begin{aligned}
    \mathop{\min}\limits_{u_0^i,u_1^i..u_{T-1}^i} & \gamma_i - \gamma_j \\
    s.t. &x_{t+1}^k=f(x_t^k,u_t^k), \forall t \in {0,1,..,T-1} k \in {i,j}\\
    &x_{t+1}^k=f(x_t^k,u_t^k), \forall t \in {0,1,..,T-1} \forall k \in {i,j}\\
    &q(x_t^k) \le w \\
    &d(x_t^i,x_t^j) \forall k \in {i,j} \ge 0
\end{aligned}
\end{equation}

This formulation is similar to \cite{Thakkar2022HierarchicalCF} except for the transition model $f$. For more details, readers are referred to it. The dynamic bicycle model is used to define the model transition $f$. The dynamic model state $s_t$ is defined with global coordinates $x$, $y$, and yaw rotation $\phi$ in the global frame; longitudinal velocity $v_x$, lateral velocity $v_y$, and yaw angular velocity $\omega$ in the vehicle's body frame. Throttle $d$ and steering $\delta$ define the action space of the model. $F_{r,x}$ is the longitudinal force on the rear tire in the tire frame assuming a rear-driven vehicle, $F_{f,y}$ and $F_{r,y}$ are the forces on the front and rear tires, respectively, and $\alpha_f$ and $\alpha_r$ are the corresponding slip angles. We denote the mass of the vehicle $m$, the moment of inertia in the vertical direction about the center of mass of the vehicle $I_z$, the length of the vehicle from the COM (center of mass) to the front wheel $l_f$, and the length from the COM to the rear wheel $l_r$. $B_{f/r}$, $C_{f/r}$, $D_{f/r}$ are the Pacejka tire model parameters specific to the tire and track surface. For longitudinal force, $C_{m1}, C_{m2}$ are known constants obtained from the gear model and $C_r, C_d$ are aerodynamic force constants which are learned from vehicle interactions. Mathematically, the vehicle model $f$ is defined as follows:

\begin{equation} \label{eqn:dyn_model}
\begin{split}
&\begin{bmatrix}
    \dot{x} \\
    \dot{y} \\
    \dot{\phi} \\
    \dot{v}_x \\
    \dot{v}_y \\
    \dot{\omega} \\
\end{bmatrix}
= \begin{bmatrix}
    v_x \cos (\phi) - v_y \sin (\phi)\\
    v_x \sin (\phi) + v_y \cos (\phi) \\
    \omega \\
    \frac{1}{m} (F_{r,x} - F_{f,y} \sin(\delta) + m v_y \omega - m g \sin(p))\\
    \frac{1}{m} (F_{r,y} + F_{f,y} \cos(\delta) - m v_x \omega + m g \sin(r))\\
    \frac{1}{I_z} (F_{f,y} l_f \cos(\delta) - F_{r,y} l_r) \\
\end{bmatrix} \\ 
\end{split}
\end{equation}
where $F_{r,x} = (C_{m1} - C_{m2} v_x) d - C_r - C_d v_x^2$, $F_{f,y} = D_f \sin(C_f \tan^{-1}(B_f \alpha_f)), \alpha_f = \delta - \tan^{-1}\left(\frac{\omega l_f + v_y}{v_x}\right)$, and $F_{r,y} = D_r \sin(C_r \tan^{-1}(B_r \alpha_r)), \alpha_r = \tan^{-1}\left(\frac{\omega l_r - v_y}{v_x}\right)$.

\section{Hierarchical control design}
\label{sec:proposed}

Similar to \cite{Thakkar2022HierarchicalCF}, we also propose a hierarchical design with a high-level planner returning a discrete checkpoint plan followed by a low-level controller to track the planned checkpoints. Having a decoupled planning approach helps to achieve long-term plans like overtaking. As discussed in detail in \cite{Thakkar2022HierarchicalCF}, directly executing reinforcement learning strategies as a single controller may not allow reliably meeting all the constraints or is not strategically optimal in the long run. 

\subsection{High level planner} \label{subsec:high_level_planner}

The high-level tactical planner approximates the general game formulation discussed earlier into a simpler discrete form. This discrete game formulation requires $2$ components.

\subsubsection{State space model}

\begin{figure}
    \centering
    \includegraphics[width=0.4\textwidth]{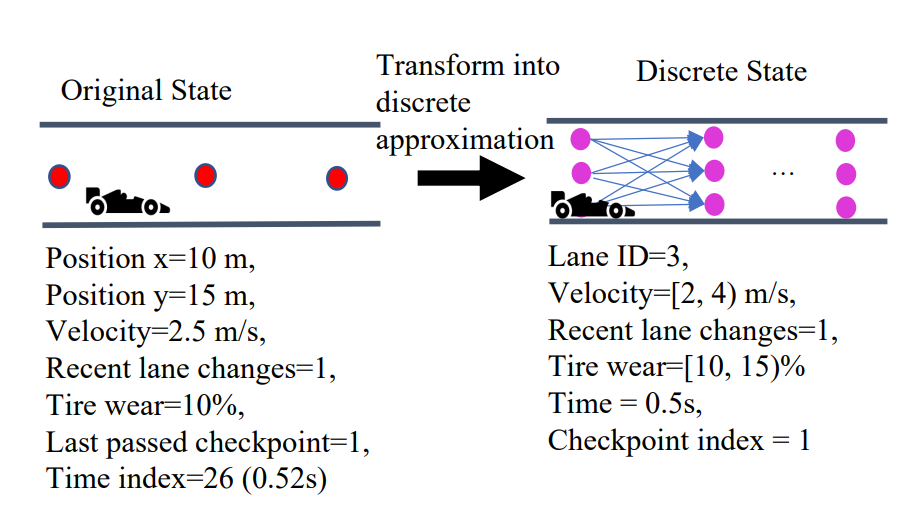}
    \caption{Converting a continuous state to discrete state}
    \label{fig:cont_to_discretel}
\end{figure}

We first transform the continuous state of the vehicle into a discrete state. We convert the position of the vehicle into a pair of discrete variables, i.e., lane ID and last passed checkpoint velocity, the latter converted to a range with suitable window size. Tire wear is also contained within a range. An example of a continuous state conversion to discrete state is shown in Fig \ref{fig:cont_to_discretel}. 

\subsubsection{Dynamics transition model}

We need a transition model to define the transition between $2$ states. If the transition is possible such that all the continuous states from the current state are able to reach at least one state on the next state, then the transition is deemed to be feasible. If all the boundary conditions are not satisfied, we rule out the transition. In our update implementation, we use simple one-dimensional equations of motion to determine current time state by taking the mean of the velocities in the range of the initial and final state \cite{Thompson1915GalileoAT}. For every state transition, the longitudinal segment is incremented strictly by $1$. The number of lane changes can be used to formulate a penalty on changing lanes too frequently on straights, which we will discuss later. 

The game is played with both players starting at an initial checkpoint and progressively updating each player's choices with the smallest time state at each point. A lower time state value implies that the player in question reached the particular checkpoint before the other player and hence gets to choose the next action. This gives the other player a chance to make a resultant strategic action, like whether and how to overtake, etc. A collision avoidance rule is incorporated in the high-level planner by restricting actions that result in the same checkpoint and the same lateral lane with time difference less than $mT$. 

\subsubsection{Solution}

\begin{figure}[htbp]
\begin{subfigure}{.24\textwidth}
    \centering
    \includegraphics[width=\textwidth]{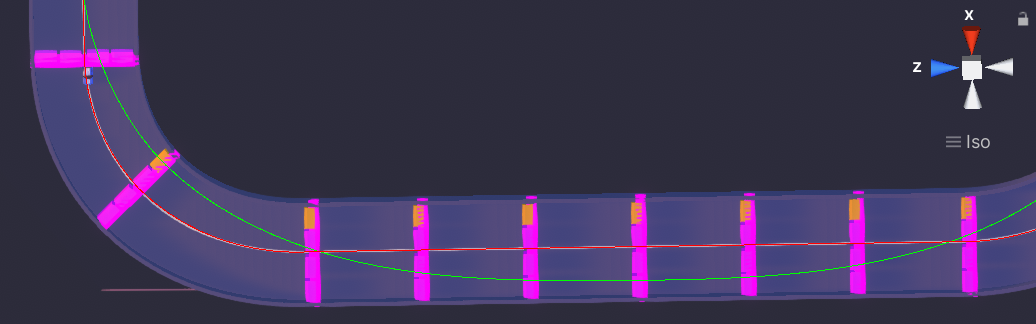}
    \caption{}
    \label{subfig:before}
\end{subfigure}
\begin{subfigure}{.24\textwidth}
    \centering
    \includegraphics[width=\textwidth]{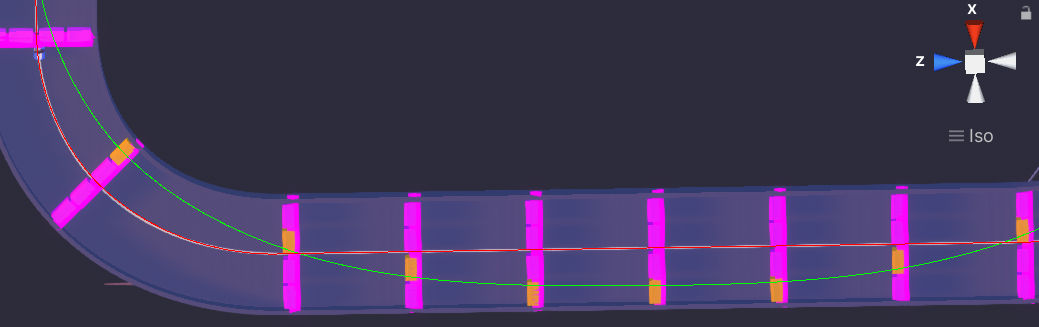}
    \caption{}
    \label{subfig:after}
\end{subfigure}
\caption{High-level plans for (a) min. time cost (b) min. distance to raceline cost}
\label{fig:with_raceline}
\end{figure}

The high-level problem of minimizing time w.r.t. the other agent is solved using monte-carlo tree search (MCTS). The optimal performance is defined as follows:

\begin{equation}
    C_{X_a,X_{a+1},..,X_{b-1}} = \sum_{i=a}^{i=b-1}(o_i-X_{i,\text{lane}})^2
\end{equation}
where $o_i$ is the optimal lane and $X_i$ is the high-level state at the $i^{\text{th}}$ checkpoint. The optimal lane is obtained from the optimal racing line. The optimal racing line is obtained from \cite{Christ2021TimeoptimalTP} by computing the time-optimal trajectory with the given vehicle model parameters. The optimal lane at each checkpoint is obtained by finding the (segment,lane) pair at which the racing line passes. The solution from MCTS is a series of discrete states both for the player and the adversarial opponent. Note that we assume that the opponent is optimal here, i.e., it too tries to achieve optimal performance. The formulation for MCTS is similar to \cite{Thakkar2022HierarchicalCF} except that the optimal solution is defined by staying closer to the racing line rather than minimizing time. We believe this leads to a better optimal solution, as explained in Section \ref{sec:results}. Figure \ref{fig:with_raceline} shows path planned from the 2 approaches. On choosing minimum time difference as the criterion for a solution, the trajectory comes out to be Figure \ref{subfig:before}, which is closer to the inner boundary, as it covers less distance, but would take a longer time in the long run. Figure \ref{subfig:after}, however, shows a new optimal trajectory along the racing line which yields shorter lap times in the long run.

\subsection{Low level controller}

The aim of the low-level controller is to execute the high-level trajectory plan. The low-level controller is typically an RL controller which takes the current state as an input and outputs the control command. 

\subsubsection{Reward design}

The reward for the low-level controller is defined by the distance to capture the following objectives:


1. Reward for passing through a checkpoint and additional reward for passing through the target lane and through the target speed : $k_{\text{target}} e^{-d_{tc}}$ where $d_{tc}$ is the distance from the target checkpoint

2. Reward for minimizing time between passing $2$ checkpoints : $-k_{\text{time}} \Delta t$ where $\Delta t$ is the time difference between passing $2$ checkpoints

3. Negative reward for swerving too frequently on straights : $-k_{\text{swerve}} \mathbb{1}_{(x,y) \in S}$ where $S$ is the set of straight section checkpoints

4. Negative reward for colliding with the wall. We use an indicator function $\mathbb{1}_{I_j\le h \cup I_j \text{ hit wall}}$ that determines if the $j^{th}$ LIDAR reading is less than $h$ and if LIDAR bounced off the wall : $-\sum_{j=1}^{9} k_{\text{wall-hit}}$ $\mathbb{1}_{I_j\le h \cup I_j \text{ hit wall}}$

5. Negative reward for collision with other players. We use the indicator function $\mathbb{1}_{I_j\le h \cup I_j \text{ opponent}}$ to check if the $j^{th}$ LIDAR reading reads hitting the opponent and we have a set $\phi$ containing all LIDAR rays that point to the front of the car for which we impose additional penalty : $-\sum_{j=1}^{9} k_{\text{opp,1}} \mathbb{1}_{I_j\le h \cup I_j \text{ opponent}} + k_{\text{opp,2}} \mathbb{1}_{I_j\le h \cup I_j \text{ opponent} \cup j \in \phi}$ 

6. Negative reward for braking unnecessarily, i.e., when speed is lower than the target window, high lateral slips : $k_{\text{brake}}\textbb{1}_{v \le v_{\text{target}} \cup d \le 0} + k_{\text{slip}} (\alpha_f^2 + \alpha_r^2)$

\subsubsection{Network architecture}

PPO RL algorithm is used to obtain the optimal policy. The neural network used for estimating the value function and the policy is a simple feedforward neural network with $8$ layers and $128$ neurons on each layer, as shown in Fig. \ref{fig:train_env}. There is a \texttt{Tanh} layer at the end to restrict the output steering and throttle to their ranges. Both the steering and throttle are obtained from the output by scaling them by their ranges. The input consists of the following values:
1. The dynamic state of the vehicle consisting of ($v_x$, $v_y$, $w$);
2. The Frenet frame state w.r.t. the racing line reference, i.e., signed lateral distance from the racing line $e_1$, relative angle w.r.t. the closest point on the racing line $e_2$;
3. Relative position of the opponent vehicle;
4. Discrete high-level target state. All the values within the range are passed by the average of the lower and upper limits
5. Raw Lidar data consisting of distances at $32$ rays cast from the extreme left of the car to the right.


\subsubsection{Training environment}

\begin{figure}
    \centering
    \includegraphics[width=0.5\textwidth]{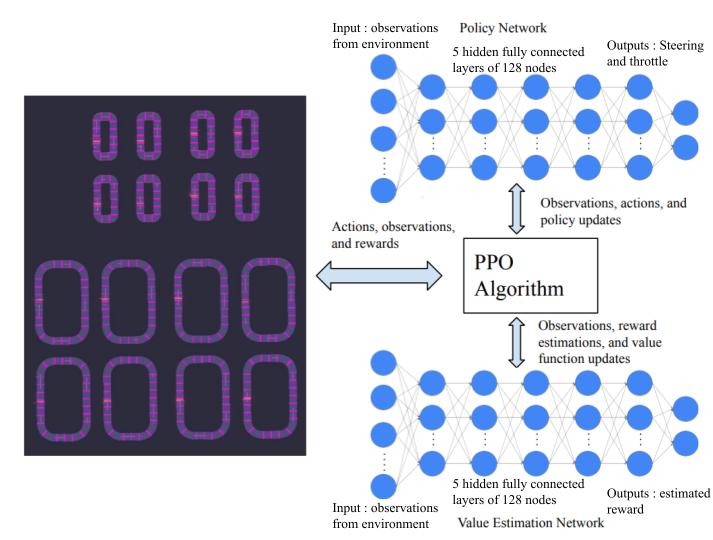}
    \caption{Training environment}
    \label{fig:train_env}
\end{figure}

Training is conducted on $16$ parallel tracks ($8$ clock-wise and $8$ counter-clockwise) so that the agent does not overfit to one side. Also, for each side, $4$ environments have steeper turns and $4$ moderate turns, as shown in Fig. \ref{fig:train_env}.

\subsection{Curriculum learning}

With the problem formulation and the hierarchical control design in place, we now define the proposed curriculum learning framework to progressively teach the RL agent an optimal policy. Let us define a parameter $t_s$ denoting the time scale. We vary $t_s$ as:

\begin{equation}
    t_s = \max \left(0,\min \left(1,\frac{t-t_{\text{start}}}{t_{\text{end}}-t_{\text{start}}} \right) \right)
\end{equation}

\subsubsection{Vehicle model transition}

The dynamic model defined in \ref{eqn:dyn_model} makes it very difficult for the RL agent to learn an effective policy, as it is very difficult to learn to move at optimal speeds while respecting the friction limits and skidding caused due to high lateral slips at higher speeds. Hence, we define a transition from a relatively simpler dynamic model which is close to the kinematic model (very few slips at the same speeds and much higher friction limits) to the complex model with the actual parameters. The tire model Pacejka parameter changes are defined as follows:

\begin{equation}
    \begin{aligned}
        &D_{ft_s} = 2^{1-t_s} D_{f0}\\
        &C_{ft_s} = C_{f0}^{2^{t_s-1}}\\
        &B_{ft_s} = 2^{1-t_s} \frac{D_{f0} C_{f0} B_{f0}}{D_{ft_s} B_{ft_s}}
    \end{aligned}
\end{equation}

\begin{figure}
    \centering
    \includegraphics[width=0.4\textwidth]{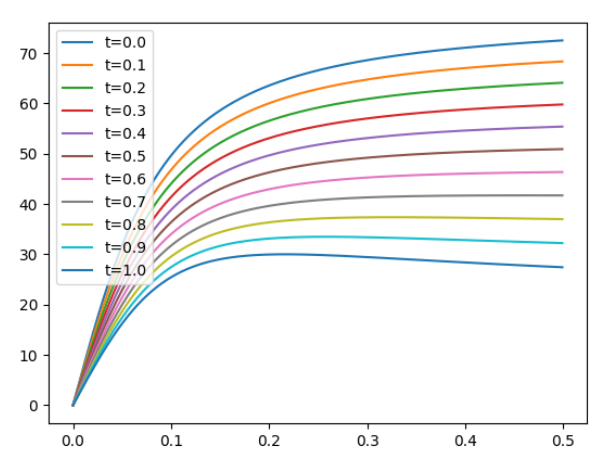}
    \caption{Tire curve variation for curriculum learning}
    \label{fig:curr_learning_tirel}
\end{figure}

\subsubsection{Safety CBF transition}

We also define a safety Control Barrier Function (CBF) to shield the agent while learning similar to \cite{Emam2021SafeRL}. We observed that the RL agent struggles a lot at the beginning hitting the walls as it tries to understand the environment. Hitting the wall once debars the agent from learning meaningful behavior later in the episode, as it gets stuck at the wall. To get rid of this, we define a safety CBF for the wall boundary constraints which overrides the RL controller by the minimum amount required to avoid hitting the wall. For more rigorous details on the CBF, readers are referred to \cite{Ames2019ControlBF}. We define the CBF function $h$ and the $2^{\text{nd}}$ order CBF as follows:

\begin{equation} \label{lane_cbf}
\begin{split}
    &h_{\text{right}}(x) = -e_{\text{center}} + w\\
    &\dot{h}_{\text{right}}(x) = - v_x\sin(\theta-\theta_{\text{ref}}) - v_y\cos(\theta-\theta_{\text{ref}})\\
    &\ddot{h}_{\text{right}}(x,u) = -\dot{v}_x \sin(\theta-\theta_{\text{ref}}) - \dot{v}_y \cos(\theta-\theta_{\text{ref}}) \\
    &\ \ \ \ \ \ \ \ \ \ \ - \omega . (v_x \cos(\theta-\theta_{\text{ref}}) -v_y \sin(\theta-\theta_{\text{ref}}))\\
    &h_{\text{left}}(x) = e_{\text{center}} + w\\
    &\dot{h}_{\text{left}}(x) = v_x\sin(\theta-\theta_{\text{ref}}) + v_y\cos(\theta-\theta_{\text{ref}})\\
    &\ddot{h}_{\text{left}}(x,u) = \dot{v}_x \sin(\theta-\theta_{\text{ref}}) + \dot{v}_y \cos(\theta-\theta_{\text{ref}}) \\
    &\ \ \ \ \ \ \ \ \ \ \ + \omega . (v_x \cos(\theta-\theta_{\text{ref}}) -v_y \sin(\theta-\theta_{\text{ref}}))\\
    &\text{Finally, }\\
    &C_{\text{right}}(x,u) = \max(0,\lambda_1 \lambda_2 \ddot{h}_{\text{right}} + (\lambda_1 + \lambda_2) \dot{h}_{\text{right}} + h_{\text{right}})\\
    &C_{\text{left}}(x,u) = \max(0,\lambda_1 \lambda_2 \ddot{h}_{\text{left}} + (\lambda_1 + \lambda_2) \dot{h}_{\text{left}} + h_{\text{left}})\\
    &\lambda_{1,t_s} = \lambda_{1,0} (1-t_s)\\
    &\lambda_{2,t_s} = \lambda_{2,0} (1-t_s)\\
\end{split}
\end{equation}

The updated command is obtained via the following optimization process where $K_{\text{viol}}$ is typically set to a very high value and $u_{\text{ref}}$ is the reference control command before the change:

\begin{equation}
    \min_{u} (K_{\text{viol}} (C_{\text{right}}^2+C_{\text{left}}^2) + |u-u_{\text{ref}}|^2)
\end{equation}

We also add a negative reward for constraint violation as follows:

\begin{equation}
    R_{\text{constraint}} = k_{\text{constraint}} (C_{\text{right}}^2+C_{\text{left}}^2)
\end{equation}

Higher values of $\lambda_1$ and $\lambda_2$ imply higher interference from the CBF, as the constraints get activated even when the agent is far from the wall, while lower values imply less interference. As the RL agent is more prone to make collisions at the beginning, higher values of $\lambda_1$ and $\lambda_2$ enable the agent to learn quickly to move along the safer centerline so as to avoid any violations. We vary the parameters of the CBF as follows:

\begin{equation} \label{lane_cbf_change}
\begin{split}
    &\lambda_{1,t_s} = \lambda_{1,0} (1-t_s)\\
    &\lambda_{2,t_s} = \lambda_{2,0} (1-t_s)\\
\end{split}
\end{equation}

\section{Results} \label{sec:results}

\begin{figure}[htbp]
\begin{subfigure}{.24\textwidth}
    \centering
    \includegraphics[width=\textwidth]{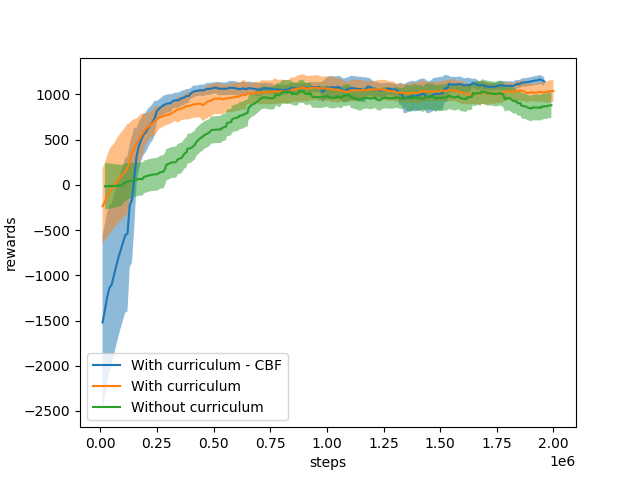}
    \caption{}
    \label{fig:rewards}
\end{subfigure}
\begin{subfigure}{.24\textwidth}
    \centering
    \includegraphics[width=\textwidth]{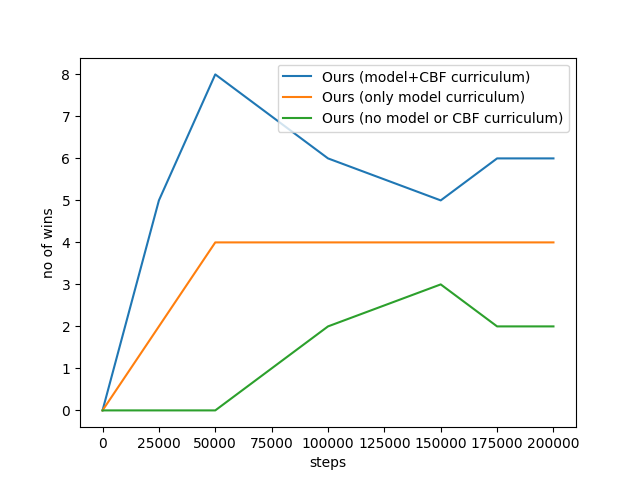}
    \caption{}
    \label{fig:n_wins_iter}
\end{subfigure}
\caption{(a) Training rewards (b) No of wins across training steps}
\end{figure}

\begin{figure}
    \centering
    \includegraphics[width=0.5\textwidth]{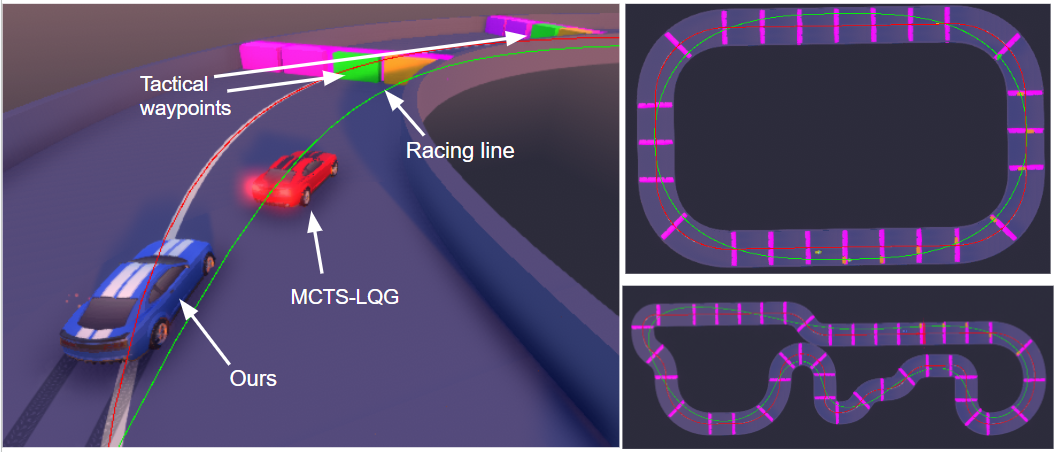}
    \caption{The race setup}
    \label{fig:race_setup}
\end{figure}

Our framework is implemented in the Unity Game Engine, with an example representative image shown in Fig. \ref{fig:race_setup}. We test our controllers on $2$ track maps, as shown in Figure \ref{fig:race_setup}. We conduct test races on this track against $2$ agents to compare with them. The races are conducted with initial position set randomly either on the left or the right at the starting line with both starting at the same longitudinal level. One position may be at an advantage if it is closer to the racing line. Hence, we randomly choose the positions with an equal chance of getting either position. We first give training rewards obtained by curriculum learning to show the advantage of using curriculum learning in Fig. \ref{fig:n_wins_iter}. As can be observed, using curriculum learning clearly with only model changes clearly beats the rewards without using it. $t_{\text{start}}$ and $t_{\text{end}}$ are chosen to be $500000$ and $1500000$. It is unfair to compare till $1500000$ steps, as the first controller runs on a simple RL environment, but after $1500000$ steps both environments are the same and our controller clearly beats the non-curriculum-based RL controller in reward. Also, the number of wins with $4$ races each between $3$ pairs of agents (so effectively $12$ races) conducted every $250000$ steps clearly shows our controller wins most races at all times. With the CBF-based curriculum added, due to negative reward for each CBF constraint violation, the reward is less at the beginning but it eventually improves at the end, achieving an even larger reward. Also, it is slightly better than only using model-based curriculum learning in terms of the number of races won.

Finally, we conduct races among other baselines for comparison. All races consist of $3$ laps, with the car which crosses the finish line first after $3$ laps winning the race. In total $20$ races are conducted with each pair for comparison.


\begin{figure}
    \centering
    \includegraphics[width=0.5\textwidth]{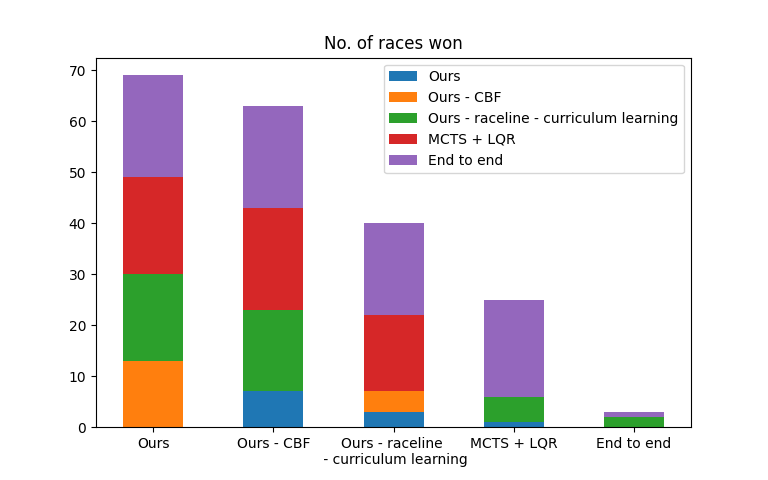}
    \caption{The race win statistics}
    \label{fig:race_win_stats}
\end{figure}

\subsection{Metrics}

We compare the runs against the following metrics: 1. No. of wins; 2. Average lap time;
3. Average lateral distance from race line; 4. No of collisions with wall; 5. No. of collisions with opponents from behind.

\begin{table}[]
\begin{tabular}{|l|l|l|ll|}
\hline
\multirow{2}{*}{}                                                               & \multirow{2}{*}{\begin{tabular}[c]{@{}l@{}}Avg. lap \\ time\end{tabular}} & \multirow{2}{*}{\begin{tabular}[c]{@{}l@{}}Avg. lateral \\ distance from \end{tabular}} & \multicolumn{2}{l|}{No of collisions} \\ \cline{4-5} 
                                                                                &                                                                           &                                                                                                    & \multicolumn{1}{l|}{Wall}  & Opponent \\ &                                                      (in s)                     &                                                              raceline (in m)                                      & \multicolumn{1}{l|}{}  &  \\ \hline
Ours                                                                            & 28.8                                                                      & 1.03                                                                                               & \multicolumn{1}{l|}{250}   & 85       \\ \hline
Ours - CBF                                                                      & 29.2                                                                      & 1.05                                                                                               & \multicolumn{1}{l|}{360}   & 92       \\ \hline
\begin{tabular}[c]{@{}l@{}}Ours - raceline, \\ curriculum learning\end{tabular} & 29.5                                                                      & 1.80                                                                                               & \multicolumn{1}{l|}{589}   & 78       \\ \hline
MCTS + LQR                                                                      & 29.6                                                                      & 0.53                                                                                               & \multicolumn{1}{l|}{212}   & 64       \\ \hline
End to end                                                                      & 30.3                                                                      & 1.67                                                                                               & \multicolumn{1}{l|}{670}   & 56       \\ \hline
\end{tabular}
\caption{Race statistics of different methods}
\label{tab:stats}
\end{table}
\subsection{Baseline methods}


We compare against the following baselines:
1. Ours 
2. Ours - CBF
3. \cite{Thakkar2022HierarchicalCF} : Ours - raceline, curriculum learning
4. MCTS + LQR : To compare against a classical rule-based controller
5. End-to-end : Not using any hierarchical controller

Fig.\ref{fig:race_setup} contains the win statistics for all methods. As can be observed, Our Method beats all other methods in most races. Especially with races against \cite{Thakkar2022HierarchicalCF}, we observe that using racing line information and our proposed curriculum learning approach is beneficial as compared to training without them. Table \ref{tab:stats} contains all the other statistics. As can be seen, we achieve better lap times as compared to other methods. Also, we stay much closer to raceline as compared to \cite{Thakkar2022HierarchicalCF} after adding explicit raceline state as input. However, using classical control like LQR is able to achieve better raceline tracking but may not be able to achieve optimal speeds as the RL controller does. Also, clearly, using the hierarchical controller yields better results than direct End-to-end approach as it looks from the average lap time. In terms of the no of collisions, it suggests that using CBF based curriculum helps the agent in learning more robust safety as it has fewer collision with wall as compared to not using CBF. 

\section{Conclusion and Future Work} \label{sec:conclusion}

In this work we propose a more realistic head-to-head racing environment to race against with more closer to actual dynamics as compared to \cite{Thakkar2022HierarchicalCF}. We then propose a hierarchical control design with a high-level controller planning a sequence of checkpoints as close to the racing line as possible and avoiding collision with other agents. We then propose a curriculum-based learning method to effectively learn an optimal policy. We compare the results with other baseline methods. It is important to note here that this is a work in progress and we admit that the experiments and inferences are incomplete (for example, comparison with an RL controller trained with constant CBF parameters should have been added to compare with \cite{Thakkar2022HierarchicalCF}). This is because we were not able to complete all experiments before the deadline. In the near work, we aim to also use trajectory prediction for the opponent agent instead of MCTS. We also aim to test against more complex environments where the agents would be allowed to take a pit stop for tire changes due to wear and more agents can be added into the game.
                                                                                                                                                                                                                                                                                                                                                                                                                                                                                                                                                                                                                                                                                                                                                                                                                                                                                                                                                                                                                                                                                                                                                                                                                                                                                                                                                                                                                                                                                                                                                                                                                                                                                                                      
\bibliographystyle{IEEEtran}
\bibliography{./IEEEfull,refs}

\end{document}